\documentclass{article}

\PassOptionsToPackage{numbers, compress}{natbib}

    \usepackage[preprint]{neurips_2019}

\usepackage[latin1]{inputenc} 
\usepackage[T1]{fontenc}    
\usepackage{hyperref}       
\usepackage{url}            
\usepackage{booktabs}       
\usepackage{amsfonts}       
\usepackage{nicefrac}       
\usepackage{microtype}      

\title{Probabilistic Decoupling of Labels in Classification}

\author{%
	Jeppe Nørregaard\thanks{\texttt{github.com/NorthGuard}} \\
	Cognitive Systems, Compute \\
	Technical University of Denmark \\
	2800 Kgs. Lyngby, Denmark \\
	\texttt{jepno@dtu.dk} 
	\And
	Lars Kai Hansen \\
	Cognitive Systems, Compute \\
	Technical University of Denmark \\
	2800 Kgs. Lyngby, Denmark \\
	\texttt{lkai@dtu.dk} 
}

\usepackage{natbib}		  
\usepackage{amsmath}
\usepackage{amssymb}
\usepackage{ mathrsfs }
\usepackage{microtype}     
\usepackage{accents}
\usepackage{setspace}		
\usepackage{bm}			    
\usepackage{dashrule}		
\usepackage{varioref}		
\usepackage{subcaption}		
\usepackage{caption}		
\usepackage{wrapfig}		
\usepackage{color}			
\usepackage[dvipsnames]{xcolor}		
\usepackage{multicol}		
\usepackage[super]{nth}		
\usepackage{bold-extra} 	
\usepackage{graphbox}		

\usepackage{longtable}		
\usepackage{stmaryrd}		
\usepackage{listings}		
\usepackage{algorithm}      
\usepackage{algpseudocode}	
\usepackage{pdfpages}		
\usepackage{sidecap}		
\sidecaptionvpos{figure}{c}
\usepackage{multirow}		
\usepackage{framed}			
\usepackage{units}			
\usepackage{bold-extra}		
\usepackage[export]{adjustbox} 
\usepackage{booktabs}		
\usepackage{needspace}		
\usepackage{relsize}		
\usepackage{pifont} 		
\usepackage{threeparttable}
\usepackage[counterclockwise, figuresleft]{rotating}		
\usepackage{upgreek}  		
\usepackage{tcolorbox}
\usepackage{setspace}
\usepackage{graphicx}

\newcommand{\diff}[2][]{\frac{\text{d}#1}{\text{d}#2}}

\newcommand{\R}{\mathbb{R}}

\DeclareMathOperator*{\mean}{mean}

\newcommand{\E}{\mathbb{E}}

\newcommand{\bone}{{\bf 1}}

\newcommand{\bn}{{\bf n}}

\newcommand{\bs}{{\bf s}}

\newcommand{\bx}{{\bf x}}

\newcommand{\bA}{{\bf A}}

\newcommand{\bC}{{\bf C}}

\newcommand{\bM}{{\bf M}}

\newcommand{\bS}{{\bf S}}
\newcommand{\bT}{{\bf T}}

\newcommand{\bW}{{\bf W}}
\newcommand{\bX}{{\bf X}}
\newcommand{\bY}{{\bf Y}}

\newcommand{\blambda}{{\boldsymbol 	\uplambda}}

\newcommand{\bLambda}{{\boldsymbol 	\Uplambda}}

\newcommand{\bUpsilon}{{\boldsymbol \Upupsilon}}

\makeatletter
\newsavebox\myboxA
\newsavebox\myboxB
\newlength\mylenA
\newcommand*\xbar[2][0.75]{%
	\sbox{\myboxA}{$\m@th#2$}%
	\setbox\myboxB\null
	\ht\myboxB=\ht\myboxA%
	\dp\myboxB=\dp\myboxA%
	\wd\myboxB=#1\wd\myboxA
	\sbox\myboxB{$\m@th\overline{\copy\myboxB}$}
	\setlength\mylenA{\the\wd\myboxA}
	\addtolength\mylenA{-\the\wd\myboxB}%
	\ifdim\wd\myboxB<\wd\myboxA%
	\rlap{\hskip 0.5\mylenA\usebox\myboxB}{\usebox\myboxA}%
	\else
	\hskip -0.5\mylenA\rlap{\usebox\myboxA}{\hskip 0.5\mylenA\usebox\myboxB}%
	\fi}
\makeatother

\newcommand{\given}{\: | \:}

\newcommand{\temp}{def}
\newcommand{\tempp}{def}
\newcommand{\temppp}{def}
\newcommand{\tempppp}{def}

\usepackage[textsize=scriptsize, textwidth=3.2cm, disable]{todonotes}		
\presetkeys{todonotes}{backgroundcolor=green, linecolor=black}{}

\usepackage{enumitem}
\DeclareRobustCommand*\circled[1]{\tikz[baseline=(char.base)]{
		\node[shape=circle,draw,inner sep=1pt] (char) {#1};}}

\begin{document}

\maketitle

\begin{abstract}
	We investigate probabilistic decoupling of labels supplied for training, from the underlying classes for prediction. Decoupling enables  an inference scheme general enough to implement many classification problems, including supervised, semi-supervised, positive-unlabelled, noisy-label and suggests a general solution to the multi-positive-unlabelled learning problem. We test the method on the Fashion MNIST and 20 News Groups datasets for performance benchmarks, where we simulate noise, partial labelling etc.
\end{abstract}

\section{Introduction}

In this work we analyse a scheme for decoupling of labels provided for training and class labels. This creates an inference scheme which can be generalized to many interesting types of classification problems, including supervised, semi-supervised, positive-unlabelled and noisy labels learning. It also provides a natural way of combining labels from different dataset. We term the labels from a dataset selection labels (or just labels) and denote them by $s \in \mathcal{S}$, where $\mathcal{S}$ is the set of labels. We use the term classes for the wanted predictions from our models, which is similarly denoted by $y \in \mathcal{Y}$.

\section{Previous Work}

The field of Noisy Label Classification concerns situations where the labels given to a machine learning algorithms have probabilistic relations to the classes, expressed as incorrect labels. In these situation it is essential to distinguish between the true \textit{class} of an instance and the given \textit{label} of an instance \cite{zhu_class_2016} - a convention we will use as well. Frenay and Verleysen \cite{frenay_classification_2014} provides a survey of noisy label classification and details a taxonomy of label noise. One characterization of noise is adhere to the Noisy at Random Model (NAR) and assumption, which assumes that the probability of error depends on the true class of a sample, but is independent on all other variables (the input space). Under the NAR model, one can characterize the labelling using a transition matrix \cite{perez_misclassified_2007}, which specifies the probability of a sample from a class getting a specific label. Bayesian approaches has been used to compute predictive distributions of the true proportions of each class, and the transition matrix, based on labels and priors on proportions and transitions \cite{perez_misclassified_2007}\cite{ruiz_bayesian_2008}. Lawrence and Sch\"{o}lkopf \cite{d_lawrence_estimating_2009} creates and EM algorithm for updating labels using Gaussian densities on classes. The E-step predicts the class-distribution for each samples and the M-step updates the parameters for each model using these distributions. Li et al.\cite{li_classification_2007} extends this method to using kernel methods, providing the new update rules for the EM-algorithm, and \cite{bootkrajang_multi_class_2010} extends to multiple classes while still using Gaussians. Some methods rely on outlier detection by training algorithms on data for evaluating the data itself, for example through cross-prediction \cite{brodley_identifying_1999}, ensembles \cite{memon_generating_2004} and iterative updates of labels \cite{zhu_class_2006}.

Semi-Supervised Learning deals with the problem of utilizing unlabelled data together with labelled data for better performance in learning algorithms \cite{sheikhpour_survey_2017}\cite{pise_survey_2008}. Kernel based methods can be used to directly estimate class densities to label unlabelled samples, if they are predicted with high confidence. They also create the foundation of semi-supervised support vector machines (S$^3$VMs)\cite{p._bennett_semi-supervised_2009}, or they can be used to impose graphs on data which can be utilized to distribute labels onto unlabelled instances\cite{zhu_learning_2009}. Many methods perform this action of evaluating unlabelled instances in order to spread out labels and extent the class distributions to unlabelled regions \cite{rosenberg_semi-supervised_2007}. 
Generative models are also widely applied in semi-supervised learning, where a generative model is trained to learn density structures in the feature space, where the labelled instances are used to label these density regions\cite{kingma_semi-supervised_2014}\cite{prescott_adams_archipelago:_2010}.

A branch of semi-supervised learning is positive-unlabelled learning, in which we have access to labelled data, which only contains a single class, and to unlabelled data which will contain the class we are looking for as well as other samples. Elkan and Noto \cite{elkan_learning_2008} decouples selection labels from classes in a similar manner to noisy labelling. In their framework, only positive samples ($y=1$) may be selected ($s=1$). They prove that if positive samples are selected at random; $p(s \given y, x) = p(s \given y)$ and if only positive samples are selected; $p(s=1 \given y=0) = 0$, then $p(y=1 \given x) = \frac{p(s=1 \given x)}{p(s=1 \given y=1)}$. Thus if you can train an algorithm to predict $p(s \given x)$ and if you can estimate $p(s=1 \given y=1)$, then you can transform you predictions into estimating $p(y=1 \given x)$. They also provide three ways of estimating $p(s=1 \given y=1)$. In \ref{sec:relation_to_elkan_and_noto_2008} in the supplementary methods we show that the methodology of \cite{elkan_learning_2008}
is a special case of the methods in this paper.

Multi-positive learning is a generalization of positive-unlabelled learning, in which there are multiple labelled positive classes and a single negative one (which may represent the joint of multiple unseen classes). Multi-positive and unlabelled learning has received limited attention despite its importance and the popularity of its special case; positive-unlabelled learning. Xu et al.\cite{xu_multi-positive_2017} derives a loss function for linear model operating in a multi-positive learning setting and proposes an iterative algorithm which switches between updating parameters of a classification model and using the model to label unlabelled samples.

\section{Decoupling Labels from Classes}

\subsection{Main Lemma}

Consider a dataset with samples from an input space collected in matrix $\bX \in \R^{n \times d}$, where $n$ is the number of samples and $d$ is the dimensionality of the input space. Each sample is selected for exactly one label and these labels are gathered in a one-hot encoded matrix
\begin{align*}
&\bS_{\mathcal{D}} \in \{0, 1\}^{n \times m_s}, \quad \bS_{\mathcal{D}} \bone = \bone,
\end{align*}
where $m_s$ is the number of possible selection labels and the $\bone$'s are vectors of ones (of suitable dimensionality). All samples have exactly one label and we will therefore have a dedicated label for "unlabelled samples" if needed. 
We assume each sample belongs to one of a set of classes, but unlike regular classification we consider the selection labels disjoint from these classes. Let the following be the unknown, true classes for the samples
\begin{align*}
&\bY_{\mathcal{D}} \in \{0, 1\}^{n \times m_y}, \qquad \bY_{\mathcal{D}} \bone = \bone,
\end{align*}
where $m_y$ is the number of classes. 

We wish to estimate the class probabilities conditioned on the input space $p(y \given \bx)$. For a set of samples we therefore define the following matrix 
\begin{align*}
&\bY \in [0, 1]^{n \times m_y}, 
\quad 0 \leq \bY_{i y} = p(y \given \bx_i), 
\quad \bY \bone = \bone.
\end{align*}

The conditional selection probabilities $p(s \given \bx)$ can be similarly collected
\begin{align*}
&\bS \in [0, 1]^{n \times m_s}, 
\quad 0 \leq \bS_{is} = p(s \given \bx_i), 
\quad \bS \bone = \bone.
\end{align*}

We will refer to the probabilities of selections conditioned on classes $p(s \given y)$ as transitions, as is customary in noisy-label learning and for similar variables in for example Markov processes. We collect the transition probabilities in a matrix as well
\begin{align*}
&\bT \in [0, 1]^{m_y \times m_s}, 
\quad 0 \leq \bT_{y s} = p(s \given y), 
\quad \bT \bone = \bone.
\end{align*}

Assume random sampling of selection labels within the classes, so that $\bx$ and $s$ are conditionally independent given $y$:
\begin{align*}
	p(s \given y, \bx) = p(s \given y).
\end{align*}
Note that in general the opposite assumption does not hold: $p(y \given s, \bx) \not= p(y \given s)$. 
Then the probability of a selection $s$ for a sample becomes
\begin{align}
p(s \given \bx) &= \sum_{y} p(s \given y, \bx) \; p(y \given \bx) \nonumber = \sum_{y} p(s \given y) \; p(y \given \bx), \label{eq:selection_from_classes}
\end{align}
which for a set of samples can be expressed as a linear equation by
\begin{align}
	\bS = \bY\bT.
\end{align}
While tempting to isolate $\bY$ using the inverse or pseudo-inverse of $\bT$, for most situations this is not a suitable approach and will usually results in negative and unscaled values (for the probabilities). We will later show an alternative approach to determine $\bY$, and possibly $\bT$ as well.

\subsection{Transition Matrix $\bT$}  \label{subsec:transition_matrix}
\renewcommand{\temp}{0.25}
\begin{figure*}[t]
	\centering
	\begin{subfigure}{0.2\textwidth}
		\centering
		\includegraphics[scale=\temp]{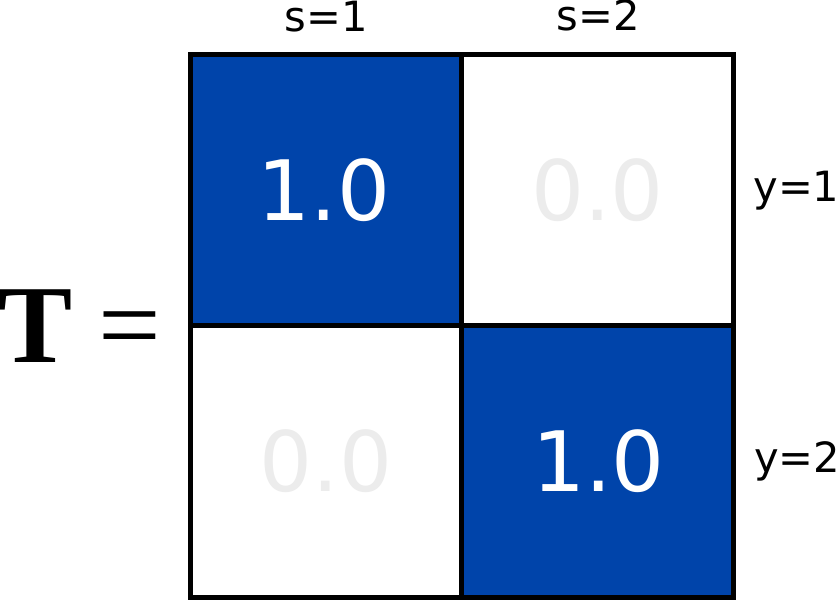}
		\caption{Binary, supervised learning.} \label{subfig:T_supervised}
	\end{subfigure}~~~~
	\begin{subfigure}{0.2\textwidth}
		\centering
		\includegraphics[scale=\temp]{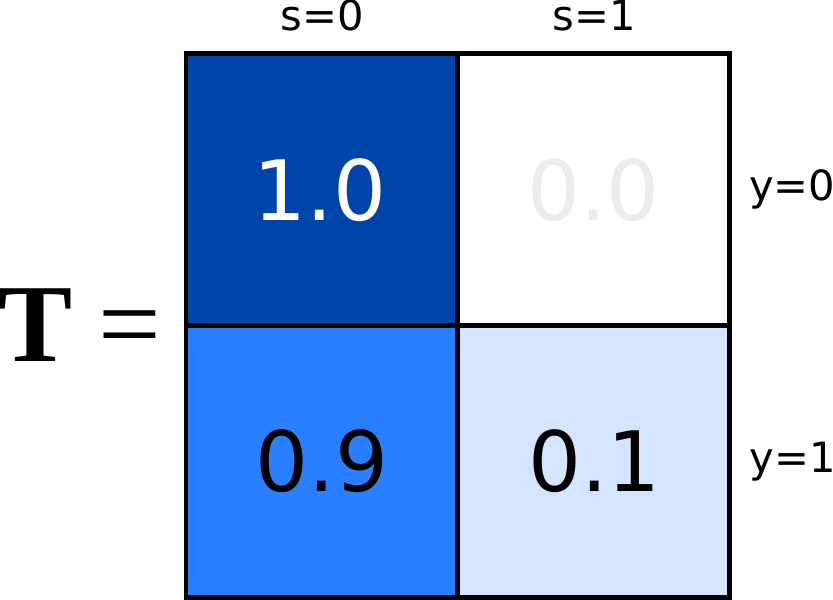}
		\caption{Positive-unlabelled learning.} \label{subfig:T_pu}
	\end{subfigure}~~~~
	\begin{subfigure}{0.2\textwidth}
		\centering
		\includegraphics[scale=\temp]{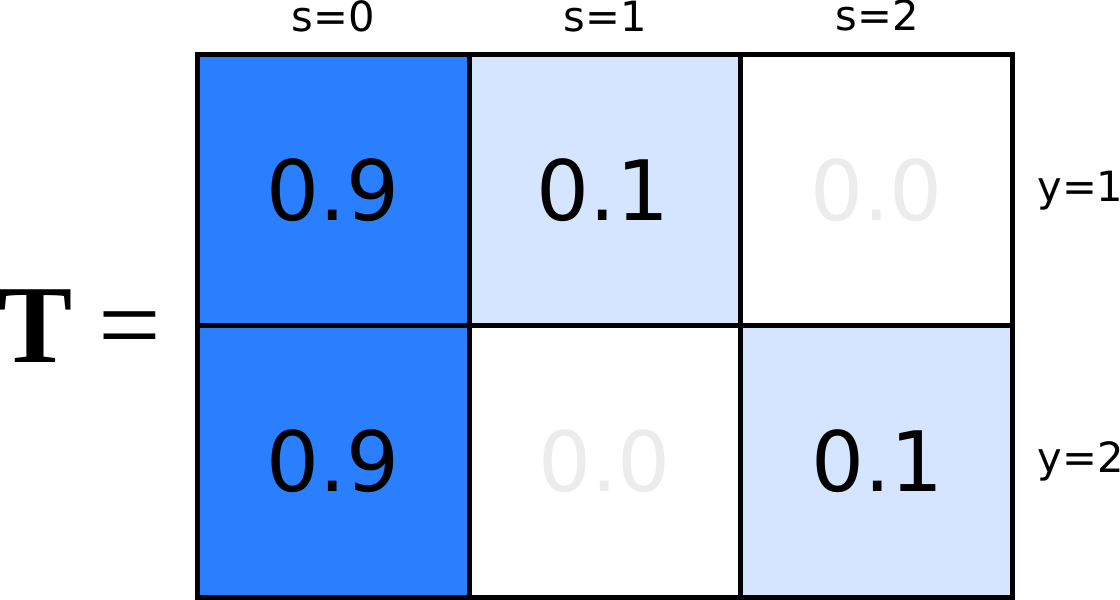}
		\caption{Binary, semi-supervised learning.}  \label{subfig:T_semi_supervised}
	\end{subfigure}~~~~
	\begin{subfigure}{0.2\textwidth}
		\centering
		\includegraphics[scale=\temp]{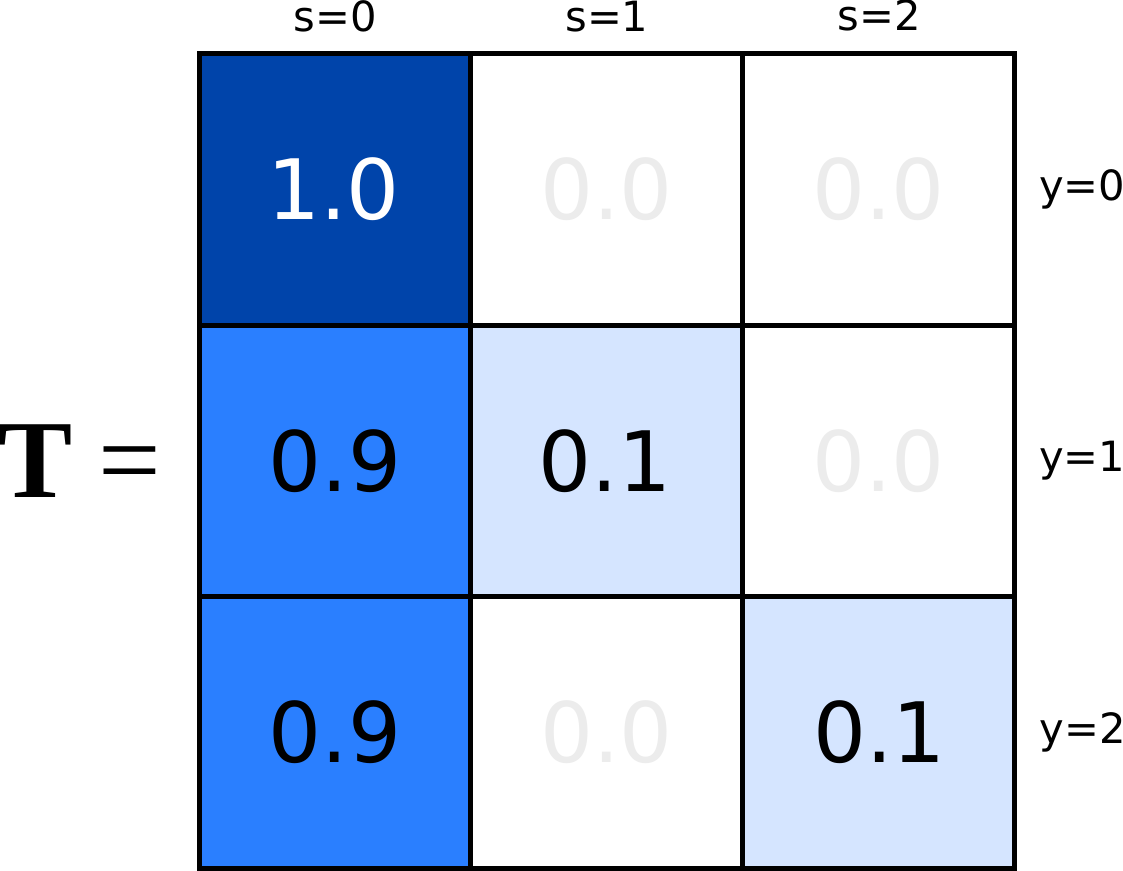}
		\caption{Semi-supervised learning with negative class.} \label{subfig:T_semi_supervised_w_negative}
	\end{subfigure}\\
	\begin{subfigure}{0.32\textwidth}
		\centering
		\includegraphics[scale=\temp]{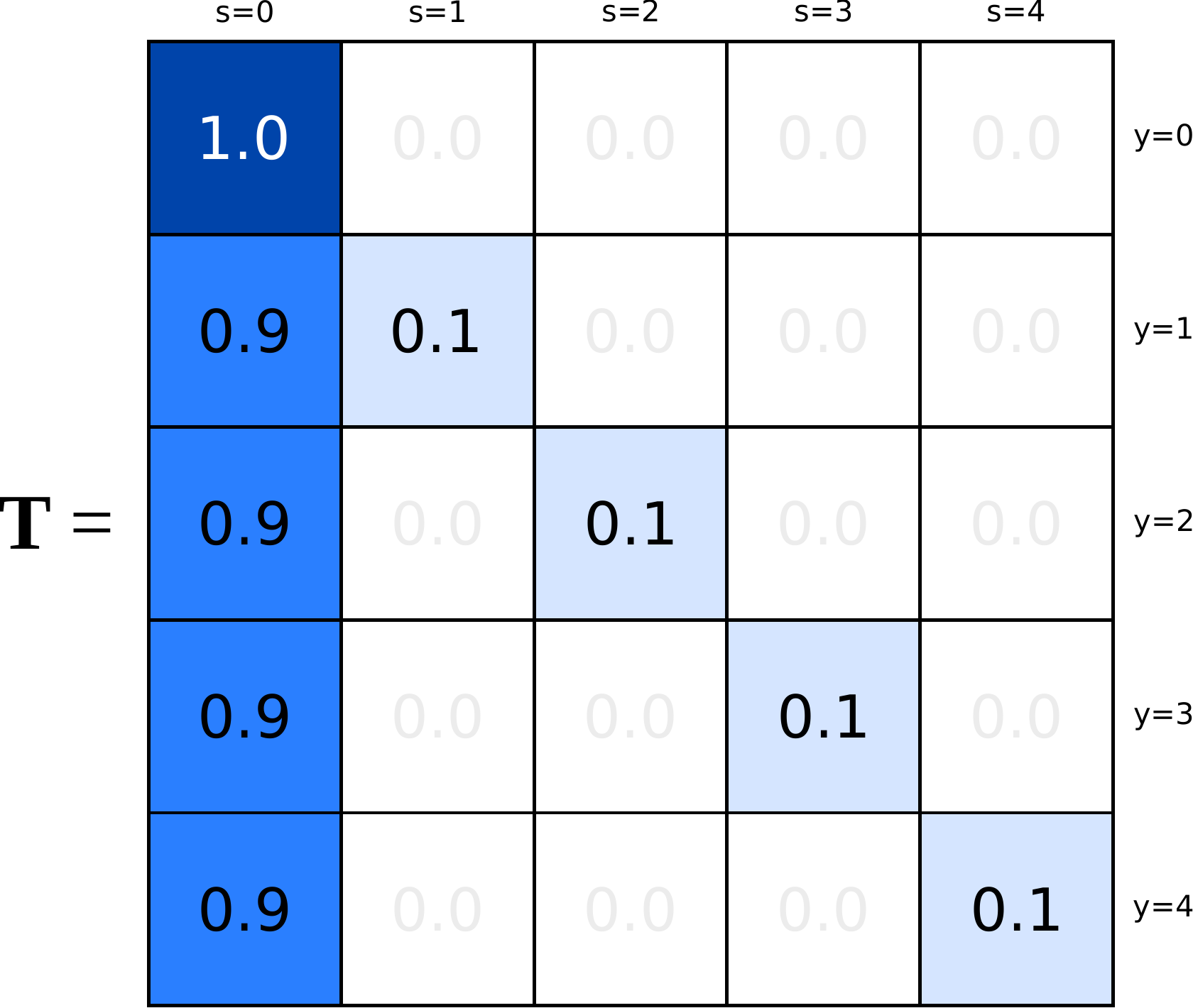}
		\caption{Semi-supervised learning with negative class and multiple positive classes.} \label{subfig:T_multipos_w_negative}
	\end{subfigure}~~~~
	\begin{subfigure}{0.32\textwidth}
		\centering
		\includegraphics[scale=\temp]{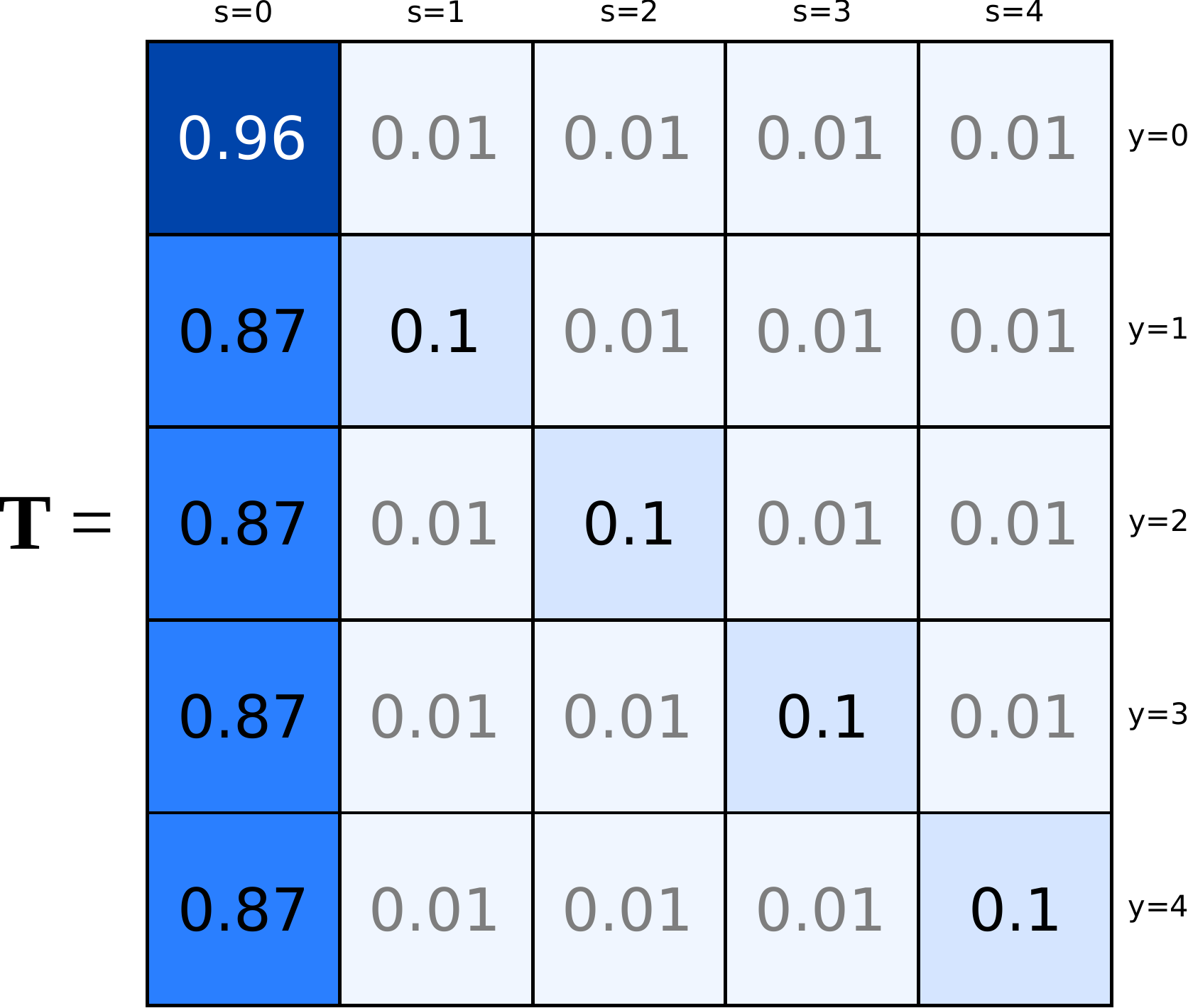}
		\caption{Semi-supervised learning with negative class, multiple positive classes and noisy labels.} \label{subfig:T_multipos_w_neg_noise}
	\end{subfigure}~~~~
	\begin{subfigure}{0.32\textwidth}
		\centering
		\includegraphics[scale=\temp]{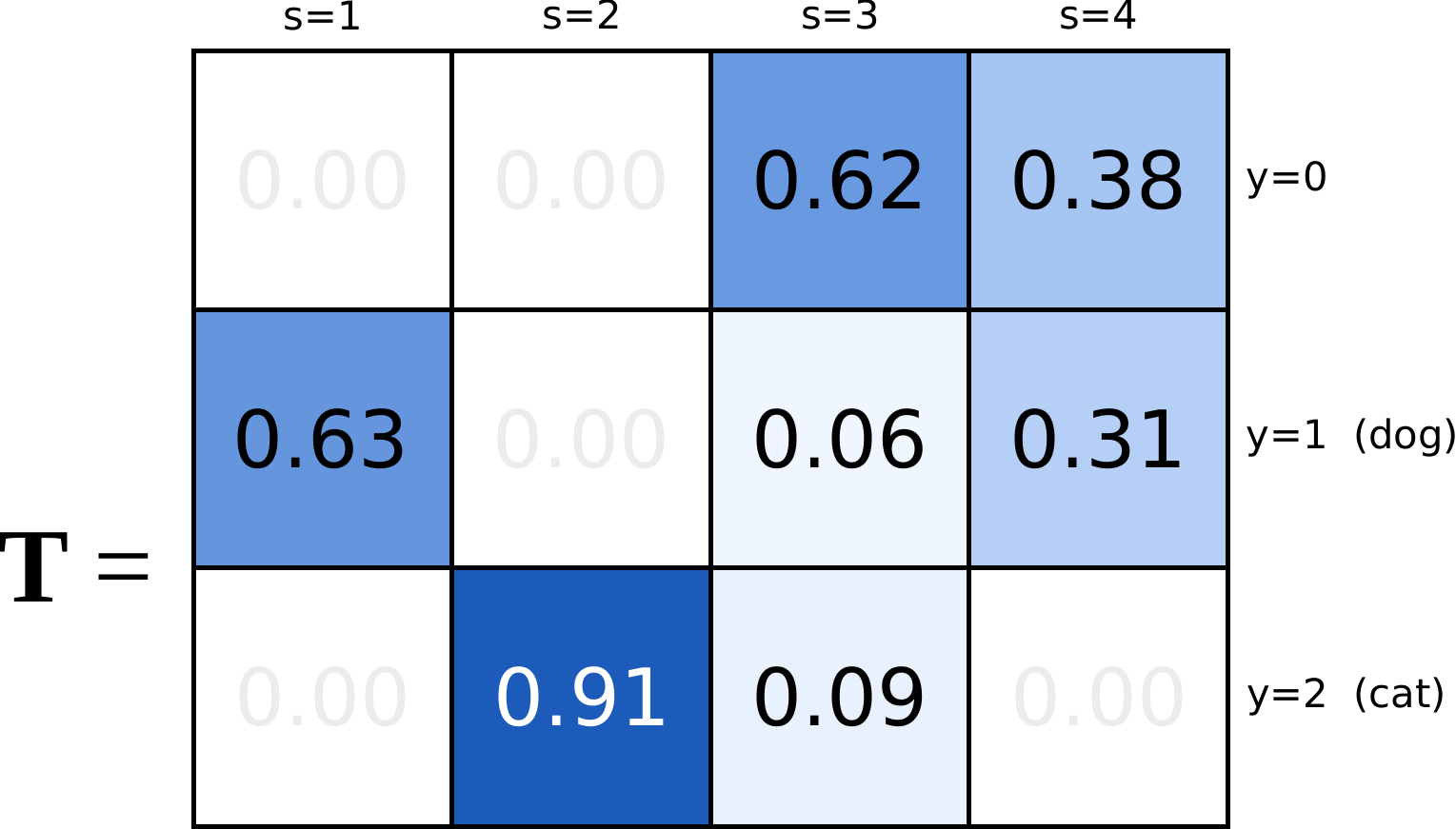}
		\caption{Semi-supervised learning with negative class and two positive classes, which exemplifies data integration using $\bT$ (section \ref{subsec:data_integration}).} \label{subfig:T_multipos_w_neg_one_to_many}
	\end{subfigure}~~~~
	\caption{Examples of $\bT$ matrices for different types of learning. The figures are explained in more detail in sections \ref{subsec:transition_matrix} and \ref{subsec:data_integration}.} \label{fig:T_examples}
\end{figure*}

Let's consider the applications of the transition matrix $\bT$. Different types of problems in classification tasks can be addressed when designing the transition matrix. We use the term "positive class" to refer to classes which we want to identify, and "negative class" to the class for all samples that we are not searching for. For example a system for detecting dogs and cats in images will have two positive classes; dog and cat, and one negative class containing all images that do not have a dog or a cat in them. There is no labels for dividing "negative classes", and thus we only consider cases with one negative class. Figure \ref{fig:T_examples} shows different transition matrices, which we will discuss here. 

Figure \ref{subfig:T_supervised} shows $\bT$ where the selection corresponds exactly to the classes in binary classification. 
Figure \ref{subfig:T_pu} shows a positive-unlabelled setting. Here $\bT$ shows that 10\% of the positive samples are labelled, while the remaining 90\% and all of the negative class are unlabelled. 
Figure \ref{subfig:T_semi_supervised} is semi-supervised case, where we assume all unlabelled data comes from one of the two positive classes. A subset of samples are labelled in the correct classes, while most of the data is unlabelled. 
Figure \ref{subfig:T_semi_supervised_w_negative} is also semi-supervised, but here we allow some unlabelled samples to come from the negative class. That is, some samples do not belong to any of our positive classes. 
(\ref{subfig:T_multipos_w_negative}) generalizes binary, semi-supervised learning to having multiple positive classes. 
Finally (\ref{subfig:T_multipos_w_neg_noise}) shows a semi-supervised, multi-positive class problem with noisy labels. We have a 1\% chance of mislabelling a sample. Of course the noise rate can differ across the matrix for the different transitions.  

If the true values for $\bY$ and $\bS$ are known we have $\bY = \bS \bT^{-1}$ (if $\bT^{-1}$ exists). Many of the elements of $\bT^{-1}$ can be negative. Actually, the inverse of a non-negative matrix is non-negative if and only if it is a scaled permutation matrix\cite{lauerberg_theorems_2008}\cite{minc_nonnegative_1988}. This will only hold if there is a one-to-one correspondence between classes and labels, which is the fully-labelled, non-noisy, multi-class classification case. The inverse $\bT^{-1}$ does thus generally not correspond to a transition matrix. The negative values creates constraints on possible values of $\bS$ to ensure non-negativity and normalization of the rows of $\bY$. While these constraints are difficult to handle analytically for large problems, in the positive-unlabelled case they can be easily interpreted and provide same constraints as those found by \cite{elkan_learning_2008}. This is shown in the supplementary material \ref{sec:relation_to_elkan_and_noto_2008}.

\subsection{Reverse Transitions} \label{subsec:inverse_T}

Let $\bUpsilon$ be the probabilities of the reverse transitions
\begin{align}
&\bUpsilon \in [0, 1]^{m_s \times m_y}, 
\quad 0 \leq \bUpsilon_{sy} = p(y \given s), 
\quad \bUpsilon \bone = \bone.
\end{align}

The class probabilities can be composed by
\begin{align}
p(y \given \bx) = \sum_s p(y \given s, \bx) p(s \given \bx),
\end{align}
but since $p(y \given s, \bx) \not= p(y \given s)$ we can not break this further down and we find that $\bT^{-1} \not= \bUpsilon$. \\

Yet we can relate $\bT$ and $\bUpsilon$ through Bayes theorem 
\begin{align}
\bT_{ys} = p(s \given y) 
&= \frac{p(y \given s) p(s)}{p(y)} 
\approx \frac{\bUpsilon_{sy} \frac{\bn_s}{n}}{ \sum_{s'} \bUpsilon_{s'y} \frac{\bn_{s'}}{n} } 
= \frac{ \bUpsilon_{sy} \bn_s}{ \sum_{s'} \bUpsilon_{s'y} \bn_{s'} } \nonumber \\
\bUpsilon_{sy} = p(y \given s) 
&= \frac{p(s \given y) p(y)}{p(s)} 
= \frac{\bT_{ys} p(y)}{ \sum_{y'} \bT_{y's} p(y') }. \label{eq:upsilon_to_T_and_back}
\end{align}
While transition probabilities can be estimated directly from data and the reverse transition probabilities, the other way requires a prior on the class distribution. Inferring the transition probabilities from $p(y \given s)$ is useful for data integration purposes (\ref{subsec:data_integration}) as well as for determining costs of error for the inference problem (\ref{sec:cost_selection_density}). 

\subsection{Data Integration} \label{subsec:data_integration}
Decoupling labels and classes is also useful for data integration purposes. Different datasets can be combined for models predicting the same classes, but handling the labelling of each datasets differently. This is illustrated with an example here.  \\
Say you wish to classify images as being of a pet dog, or a pet cat or neither (negative class). Say we have three datasets. $\mathcal{D}_1$ is labelled with dogs ($s_1$) and cats ($s_2$) and nothing else. $\mathcal{D}_2$ is big, unlabelled, and has dogs and cats, but also other things. We give it label $s_3$. Finally we have $\mathcal{D}_3$ with label $s_4$ of canines, which will include pet dogs but no cats. Say the size ratios are $p(\mathcal{D}_1)=0.1$, $p(\mathcal{D}_2)=0.6$ and $p(\mathcal{D}_3)=0.3$. We can construct $\bUpsilon$ matrix (with example values) and determine $\bT_{\text{ex}}$ with equation (\ref{eq:upsilon_to_T_and_back}) so that
\begin{align*}
\bUpsilon_{\text{ex}} &= 
\begin{array}{l}
{\scalebox{0.7}{$s_1$}} \\ {\scalebox{0.7}{$s_2$}} \\ {\scalebox{0.7}{$s_3$}} \\ {\scalebox{0.7}{$s_4$}}
\end{array} \hspace{-2mm}
\stackrel{\text{\tiny-\hspace{6mm}dog\hspace{6mm}cat}}
{\begin{bmatrix}
	0.0 & 1.0 & 0.0 \\
	0.0 & 0.0 & 1.0 \\
	0.8 & 0.1 & 0.1 \\
	0.5 & 0.5 & 0.0 \\
	\end{bmatrix}}\hspace{-1.5mm}.
\qquad \qquad
\bT_{\text{ex}} = 
\begin{array}{r}
{\scalebox{0.7}{-}} \\ {\scalebox{0.7}{dog}} \\ {\scalebox{0.7}{cat}} 
\end{array} \hspace{-5mm}
\stackrel{\text{
		\hspace{6mm}
		\scalebox{0.9}{$s_1$}\hspace{6.6mm}
		\scalebox{0.9}{$s_2$}\hspace{6.6mm}
		\scalebox{0.9}{$s_3$}\hspace{6.6mm}
		\scalebox{0.9}{$s_4$}\hspace{6.6mm}
}}
{\begin{bmatrix}
	0.00 & 0.00 & 0.62 & 0.38 \\
	0.63 & 0.00 & 0.06 & 0.31 \\
	0.00 & 0.91 & 0.09 & 0.00 \\
	\end{bmatrix}}\hspace{-4mm}.
\end{align*}

This transition matrix can be used to train models for the classes instead of the selections, while handling the differences in labelling methodologies. When transitions are directly known instead of the reverse transitions, then these can of cause be used instead. Figure \ref{subfig:T_multipos_w_neg_one_to_many} shows $\bT_{\text{ex}}$.

\section{Inference} \label{sec:inference_Y_T}

We wish to learn a function $f(\bx) \approx p(y \given \bx)$ from a set of samples and selection labels. Say we train an alternative classification function to estimate the probabilities of selection instead $g(\bx) \approx p(s \given \bx)$. Using this approximation we can create an approximated selection matrix $\widehat{\bS}$, which we wish to use to infer an approximated class probability matrix through
\begin{align*}
\widehat{\bS} = \widehat{\bY} \bT.
\end{align*}
In the following we show how to infer $\widehat{\bY}$ using $\widehat{\bS}$ and a known $\bT$, and even how to estimate $\widehat{\bT} \approx \bT$ simultaneously using $\widehat{\bS}$ and a prior on $\bT$. An important note on this problem is that this inference problem has multiple solutions due to permutations; we can switch the columns of $\widehat{\bY}$ and rows of $\widehat{\bT}$ resulting in solutions with equal likelihood. In this paper we handle the permutation problem using priors, but it can be problematic if little is known about the classes.

We are also interested in determining the class probabilities of our labelled dataset, which is non-trivial due to unlabelled samples and noisy labels. We therefore define $\bW$ be the class distributions of the samples conditioned on the selection
\begin{align*}
&\bW \in [0, 1]^{n \times m_y}, 
\quad \bW_{iy} = p(y \given \bx_i, \bS_{\mathcal{D}}), 
\quad \bW \bone = \bone.
\end{align*}
We consider the following tasks \\[3pt]
\begin{minipage}{\textwidth}
	\begin{multicols}{2}
		\begin{enumerate}[label=\circled{\arabic*}]
			\item Infer $\widehat{\bY}$ and $\widehat{\bT}$ from $\widehat{\bS}$. \label{task:infer_Y_T_from_S}
			\item Estimate $\widehat{\bT}$ using known $\widehat{\bY}$ and $\bS_{\mathcal{D}}$. \label{task:est_T_from_Y_SD}
			\item Predict $\widehat{\bY}$ using $\widehat{\bT}$ and $\widehat{\bS}$. \label{task:predict_infer_Y_from_T_S}
			\item Estimate $\widehat{\bW}$ from $\widehat{\bY}$, $\widehat{\bT}$ and $\bS_{\mathcal{D}}$. \label{task:est_W_from_Y_T_SD}
		\end{enumerate}
	\end{multicols}
\end{minipage} \vspace{2mm}

Task \ref{task:infer_Y_T_from_S} and \ref{task:est_T_from_Y_SD} are used for inferring parameters related to a problem, task \ref{task:predict_infer_Y_from_T_S} is used to predict on new samples using these parameters, and \ref{task:est_W_from_Y_T_SD} is used to update belief about samples with selection labels (for example training samples).

\subsection{Class Densities $\bY$ and Transitions $\bT$} \label{subsec:infer_Y_T}
We now assume that we have an estimate of $p(s \given \bx)$, and we will use that to estimate $p(y \given \bx)$ and $p(s \given y)$ (task \ref{task:infer_Y_T_from_S}). The probability of selection for a sample is 
\begin{align*}
p( s \given \bx_i ) = \sum_y p(s \given y) p(y \given \bx_i) = \big( \bY \bT \big)_{is}
\end{align*}
By assigning (conjugate) Dirichlet priors for the rows of $\bT$, we can optimize $\bT$ and $\bY$ by maximizing the following optimization function
\begin{align*}
	\mathscr{O} &= - \mathcal{KL}\left( \widehat{\bS}_{is}, \widehat{\bY} \bT \right) + \sum_{iy} \bY_{iy} \log p(y) + \sum_{ys} \bA_{ys} \log \bT_{ys}. 
\end{align*}
Here $\mathcal{KL}(\cdot)$ denotes the Kullback-Leibler divergence. If we know $\bT$ then we can remove the last term in $\mathscr{O}$ and simply optimize for the optimal $\widehat{\bY}$ which solves task \ref{task:predict_infer_Y_from_T_S}. A detailed derivation of $\mathscr{O}$ can be found in \ref{sec:posterior}. Inferring $\bY$ and $\bT$ does not use the features space and each step of of an optimizer can be done in time complexity $\text{O}(m_s m_y n)$.

\subsection{Transitions $\bT$ Directly}
For observed class densities $\widehat{\bY}$ and selection $\bS_{\mathcal{D}}$, using conjugate prior in $\bT$, we can determine the aggregated mass transitioned from each class to each selection $\bM$ and the maximum a posteriori solution of $\bT$ by
\begin{align*}
\bM = \widehat{\bY}^{\top} \bS_{\mathcal{D}}, \qquad \bT_{ys} = \frac{(\bM + \bA)_{ys}}{\sum_s (\bM + \bA)_{ys}},
\end{align*}
which solves task \ref{task:est_T_from_Y_SD}. The derivation of this result can be found in the supplementary material \ref{sec:supplement_est_T}.

\subsection{Conditional Class Densities $\bW$}

We now focus on task \ref{task:est_W_from_Y_T_SD} of estimating $\bW$, which holds $p(y \given \bx, s)$ for a set of samples. Applying Bayes theorem with the class probabilities provides (using the assumption of conditional independence of $s$ and $\bx$ given $y$)
\begin{align*}
p(y \given s, \bx_i) &= \frac{p(s \given y, \bx_i) p(y \given \bx_i)}{p(s \given \bx_i)}
= \frac{p(s \given y)p(y \given \bx_i)}{ \sum_{y'} p(s \given y')p(y' \given \bx_i)} 
= \frac{\bT_{ys} \bY_{iy}}{ \sum_{y'} \bT_{y's} \bY_{iy'}}.
\end{align*}
We can pick out the relevant transitions by $\bS_D \bT^{\top}$ and estimate the probabilities for each sample by ($\circledcirc$ is the Hadamard (elementwise) product)
\begin{align*}
\widehat{\bW}_{iy} &= \frac{
	\big( (\bS_{\mathcal{D}} \widehat{\bT}^{\top}) \circledcirc \widehat{\bY} \big)_{iy}
}{
	\sum_{y'} \big( (\bS_{\mathcal{D}} \widehat{\bT}^{\top}) \circledcirc \widehat{\bY} \big)_{iy'}
} \approx p(y \given \bx_i, \bS_{\mathcal{D}}).
\end{align*}

\subsection{Costs} \label{sec:costs}
The classifier $g(\bx)$ is trained to estimate $p(s \given \bx)$ in order to later infer information about $p(y \given \bx)$. Some labels may carry more information about $y$ than others (for example the "unlabeled"-label in semi-supervised does not carry much information). In order to make our classifier carry as much information about $y$ as we can, we weight the errors made by the classifier during training. In \ref{sec:cost_selection_density} we show that this can be expressed as a cost-sensitive classification problem, with the cost of classifying a sample with label $s$ as $s'$ instead
\begin{align*}
	\bC_{ss'} = \sum_y (1 - p(y \given s')) \: p(y \given s),
	 \qquad \bC = \bUpsilon (1 - \bUpsilon)^{\top}.
\end{align*}
The error of classifying as $s'$ instead of $s$ depends on how similar the two labels are in predicting $y$. 

In practise we don't use costs between the labels, but instead weight the samples according to their selection label. We compute the weight as the expected increase in cost from selecting a different label according to the prior on the labels.

\section{Experiments}

\subsection{Simulated Data}
We first test the described method on some simulated data to show how it works. We define four 2D Gaussians (figure \ref{subfig:true_gaussians}) which we sample from. We let one of the Gaussians be the negative class. We make four selection labels, where three of them corresponds to the three positive-class Gaussians (with noisy labels) and the last label represents unlabelled. Most of the samples are unlabelled and three of the samples are labelled incorrectly. If we use a kernel density estimator for $g(\bx)$ we find the densities visualized in figure \ref{subfig:kernel_densities} together with the samples. Using the prior (almost uniform) of the labels we can estimate $p(s \given \bx)$ and we show the posterior decision regions of selection labels in figure \ref{subfig:kernel_posterior}. In figure \ref{subfig:first_inference} we have estimated $p(y \given \bx)$, $p(s \given y)$ and $p(y \given \bx, \bs)$ ($\widehat{\bY}$, $\widehat{\bT}$ and $\widehat{\bW}$). We have indicated the transition for label-domain to class-domain by a change in colours (even though for this problem there is a one-to-one correspondence between labels and classes). The regions of the classes have taken up some of the unlabelled space, but the noisy labels still cause trouble.

\renewcommand{\temp}{\textwidth}
\renewcommand{\tempp}{0.43\textwidth}
\begin{figure*}[t]
	\centering
	\begin{subfigure}{0.5\textwidth}
		\centering
		\includegraphics[width=0.72\textwidth,trim={10mm 10mm 0mm 10mm},page=9]{figs/article_plots.pdf} 
	\end{subfigure}%
	\begin{subfigure}{0.5\textwidth}
		\centering
		\includegraphics[width=0.68\textwidth,trim={10mm 10mm 0mm 10mm},page=8]{figs/article_plots.pdf} 
	\end{subfigure}\vspace{1mm}
	\begin{subfigure}{\tempp}
		\centering
		\includegraphics[width=\temp,page=1]{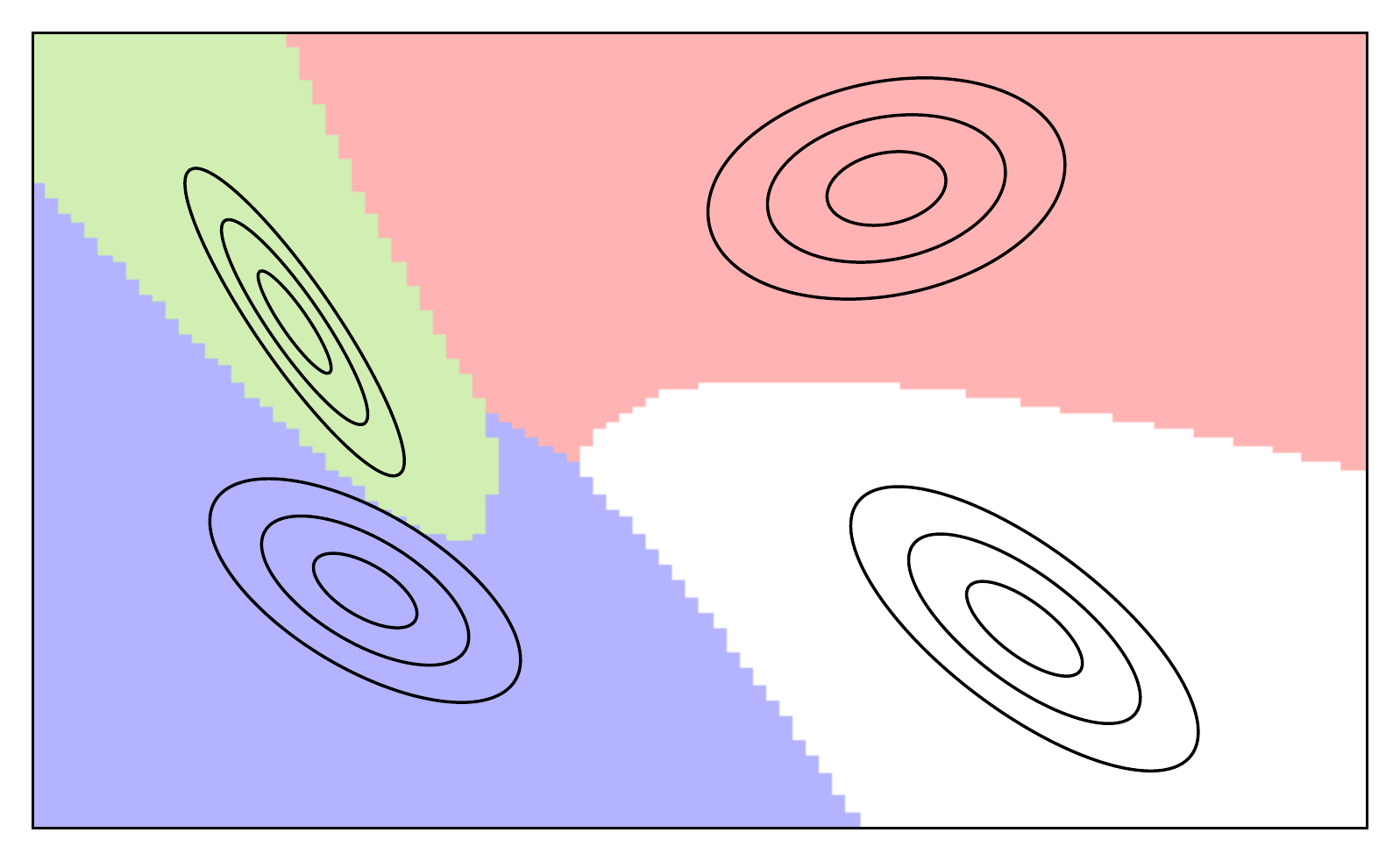}
		\caption{Four normals with posterior decision regions.} \label{subfig:true_gaussians}
	\end{subfigure}\hfil
	%
	%
	\begin{subfigure}{\tempp}
		\centering
		\includegraphics[width=\temp,page=3]{figs/article_plots.pdf}
		\caption{Samples from normals with kernel densities on selected points.} \label{subfig:kernel_densities}
	\end{subfigure}
	\begin{subfigure}{\tempp}
		\centering
		\includegraphics[width=\temp,page=4]{figs/article_plots.pdf}
		\caption{MAP decision regions for selection labels ($p(s \given \bx)$) based on kernels from figure \ref{subfig:kernel_densities}.} \label{subfig:kernel_posterior}
	\end{subfigure}\hfil
	\begin{subfigure}{\tempp}
		\centering
		\includegraphics[width=\temp,page=5]{figs/article_plots.pdf}
		\caption{MAP decision regions for classes after inference of $\widehat{\bY}$ and $\widehat{\bT}$. Colour shading is based on $p(y \given \bx)$ ($\widehat{\bY}$), while samples are coloured using $p(y \given \bx, s)$ ($\widehat{\bW}$).} \label{subfig:first_inference}
	\end{subfigure}
	\caption{Inference of $\bY$, $\bT$ and $\bW$ on simulated data. Orange, turquoise and purple indicate selection labels. Red, green and blue indicate classes.} \label{fig:simulated_data}
\end{figure*}

\subsection{Fashion MNIST} \label{sec:result_fashion}
Fashion MNIST \cite{xiao2017/online} is a dataset created by Zalando intending to be a drop-in replacement of the original MNIST dataset. They hope the dataset is more challenging and more representative of modern machine learning problems, and can thus be used as a more modern dataset for benchmarking. The dataset contains 60.000 training and 10.000 test samples of $28\times28$ grayscale images. The images are labelled into 10 classes of clothing items: T-shirt/top, trouser, pullover, dress, coat, sandal, shirt, sneaker, bag and ankle boot. The dataset furthermore contains a predetermined training-test split, which makes comparison across methods easier. We use this dataset to test our method. 


We tested the inference method by training a random forest on the training data for estimating $\widehat{\bS} \approx p(s \given \bx)$, using pixel values as features after normalizing to zero mean and unit variance. We use cross prediction in order to get unbiased predictions on the training data. After estimating the selection-densities we infer $\widehat{\bT}$ and $\widehat{\bY}$ for the training and test set, as well as $\widehat{\bW}$ for the training set.

We create the following problems while varying the amount of labelled data 
\begin{itemize}[itemsep=0pt, topsep=0pt]
	\item Semi-Supervised: predict all classes 
	\item 7-Positive: first 7 classes used as positive classes, the other 3 as negative
	\item Positive Unlabelled: first class used as positive class, the other 9 as negative
	\item Noisy-20: Like semi-supervised but with 20\% noise on labels
	\item Noisy-50: Like semi-supervised but with 50\% noise on labels
\end{itemize}
The performance of the method is shown in figure \ref{fig:results_fmnist}, with F1-score on the y-axes and number of labelled samples in training set on the x-axes. In plain semi-supervised learning (\ref{fig:semi_super_test}) the inference method does not improve performance on the test set as the unlabelled data is never really used and all information is in the labelled samples. In the 7-positives problem (\ref{fig:7_pos_test}) we see that the cost-weighted method is useful when very few samples are available and that the inference method improves performance when little data is available. When using flat weights on the samples the baseline performs quite poorly, but the inference method boosts performance quite significantly. 

For the positive-unlabelled problem the inference method heavily improves performance on the unweighted system (\ref{fig:pos_unl_test}), while the cost-weighted system has performance so high that the inference scheme can not improve it. 
For the semi-supervised, 7-positive-unlabelled and positive-unlabelled problems we can integrate the labels from the training data to compute $\bW$ which obviously improves performance as seen in figures \ref{fig:semi_super_w}, \ref{fig:7_pos_w} and \ref{fig:pos_unl_w}.

For the noisy-label problems the inference method only performs as well as the underlying classifier on the test set (thus not shown), like in the semi-supervised case. The use of $\bW$ is interesting for evaluating the noisy labels of the training data. Figures \ref{fig:noise_20_w} and \ref{fig:noise_50_w} shows the performance of using $\bW$ for the two noise-ratios. The black curve on these plots shows training-set performance, if we predict on the unlabelled set and leave the labelled set with their labels. These black curves show how the performance, when relying on the labels, degrades below what the classifier has learned, indicating that the model manages to learn the underlying distributions despite the noise on labels. Using the inference method with $\bW$ is here a much better way to use the labels and performance increases with more labels.

We also tested kernel density estimation for predicting $p(s \given \bx)$ on the dataset (\ref{sec:kernel_fashion}). This provides lower performance in general due to the restrictions of this model. We see similar results, although the performance improvements of cost-weighting and inference differ a bit across the problems. 
Furthermore we tested the system on the 20 Newsgroups dataset\cite{mitchell_20_1999} where we again used a random forest, this time on term-counts after removing stopwords. Similarly to the Fashion MNIST dataset, the 20 Newsgroups dataset have a dedicated test and training set for easy comparison. The tests showed similar results although with a general lower performance due to the difficulty of the problem. The results are shown in the supplementary material\ref{sec:20_news_groups}. 

In conclusion, we have shown how to convert a classifier predicting training labels into a classifier predicting the underlying classes. Relative to using the underlying classifier directly, this method provides significant performance increase on problems with a negative class, while providing an effective way to utilize given labels with the classifier for evaluating labels of training data.

\renewcommand{\temppp}{-2.3mm}
\renewcommand{\tempp}{10.63mm} 
\renewcommand{\temp}{0.265}
\renewcommand{\tempppp}{2mm}
\begin{figure}[t]
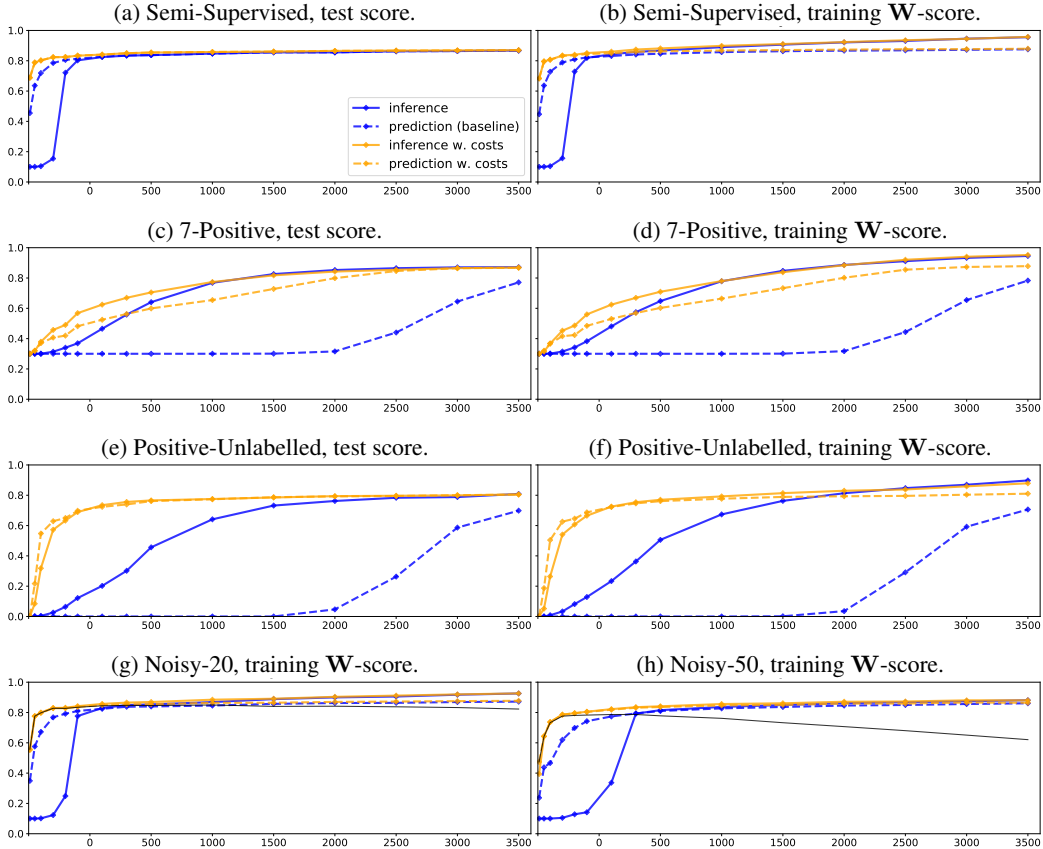

	\centering
	\begin{subfigure}{0.5\textwidth}
		\centering
		\caption{Semi-Supervised, test score.} \vspace{\temppp} \label{fig:semi_super_test}
		\includegraphics[scale=\temp,trim={0mm 0mm 0mm \tempp},page=3,clip]{figs/results/forest_results_fashion.pdf} 
	\end{subfigure}%
	\begin{subfigure}{0.5\textwidth}
		\centering
		\caption{Semi-Supervised, training $\bW$-score.} \vspace{\temppp} \label{fig:semi_super_w}
		\includegraphics[scale=\temp,trim={11mm 0mm 0mm \tempp},page=2,clip]{figs/results/forest_results_fashion.pdf} 
	\end{subfigure} \\[\tempppp]
	\begin{subfigure}{0.5\textwidth}
		\centering
		\caption{7-Positive, test score.} \vspace{\temppp} \label{fig:7_pos_test}
		\includegraphics[scale=\temp,trim={0mm 0mm 0mm \tempp},page=6,clip]{figs/results/forest_results_fashion.pdf} 
	\end{subfigure}%
	\begin{subfigure}{0.5\textwidth}
		\centering
		\caption{7-Positive, training $\bW$-score.} \vspace{\temppp} \label{fig:7_pos_w}
		\includegraphics[scale=\temp,trim={11mm 0mm 0mm \tempp},page=5,clip]{figs/results/forest_results_fashion.pdf} 
	\end{subfigure} \\[\tempppp]
	\begin{subfigure}{0.5\textwidth}
		\centering
		\caption{Positive-Unlabelled, test score.} \vspace{\temppp} \label{fig:pos_unl_test}
		\includegraphics[scale=\temp,trim={0mm 0mm 0mm \tempp},page=9,clip]{figs/results/forest_results_fashion.pdf} 
	\end{subfigure}%
	\begin{subfigure}{0.5\textwidth}
		\centering
		\caption{Positive-Unlabelled, training $\bW$-score.} \vspace{\temppp} \label{fig:pos_unl_w}
		\includegraphics[scale=\temp,trim={11mm 0mm 0mm \tempp},page=8,clip]{figs/results/forest_results_fashion.pdf} 
	\end{subfigure} \\[\tempppp]
	\begin{subfigure}{0.5\textwidth}
		\centering
		\caption{Noisy-20, training $\bW$-score.} \vspace{\temppp} \label{fig:noise_20_w}
		\includegraphics[scale=\temp,trim={0mm 0mm 0mm \tempp},page=11,clip]{figs/results/forest_results_fashion.pdf} 
	\end{subfigure}%
	\begin{subfigure}{0.5\textwidth}
		\centering
		\caption{Noisy-50, training $\bW$-score.} \vspace{\temppp} \label{fig:noise_50_w}
		\includegraphics[scale=\temp,trim={11mm 0mm 0mm \tempp},page=14,clip]{figs/results/forest_results_fashion.pdf} 
	\end{subfigure} 
	\caption{Performance on Fashion MNIST dataset. Y-axis shows F1-score on all plots, while the X-axis shows the number of labelled instances in each (positive) class. The blue curves uses flat cost on samples and the yellow curves use the costs from \ref{sec:costs}. The dashed lines are the baseline performance from predicting label with model and choosing the related class. The solid lines are the performance after inferring $\bT$ and $\bY$. Some of the plots show the $\bW$ predictions to illustrate performance on evaluating training samples. On the noise-label plot we have also included the performance gained from predicting on the unlabelled set and using the labels directly on the labelled instances (in black).} \label{fig:results_fmnist}
\end{figure}

\bibliography{references}  

\clearpage
\appendix
\section{Supplementary Material}

\subsection{Posterior for Inference} \label{sec:posterior}
We now assume that we have an estimation of $p(s \given \bx)$, and we will use that to estimate $p(y \given \bx)$ and $p(s \given y)$ (task \ref{task:infer_Y_T_from_S}). The probability of selection for a sample is 
\begin{align*}
p( s \given \bx_i ) = \sum_y p(s \given y) p(y \given \bx_i) = \big( \bY \bT \big)_{is}
\end{align*}
We observe selection probabilities $\widehat{\bS}$. The likelihood of $\bY$ and $\bT$ becomes (\ref{sec:limit_mean_likelihood})
\begin{align*}
\mathscr{L} &= p(\widehat{\bS} \given \bY, \bT, \bX) = \prod_{is} p(s \given \bx_i)^{\widehat{\bS}_{is}} 
= \prod_{is} \Big( \sum_y p(s \given y) p(y \given \bx_i) \Big)^{\widehat{\bS}_{is}}
= \prod_{is} \big( \bY \bT \big)_{is}^{\widehat{\bS}_{is}}.
\end{align*}
Assuming independence between samples we can construct the following priors on $\bY$
\begin{align*}
p(\bY) = \prod_{iy} p(y)^{\bY_{iy}}.
\end{align*}
For the transition matrix $\bT$ we assign a Dirichlet prior $\mathscr{D}$ on each row
\begin{align*}
p(\bT) = \prod_{y} \mathscr{D}_y(\bT_{y*}) = \frac{1}{B(\bA)} \prod_{ys} \bT_{ys}^{\bA_{ys}},
\end{align*}
where $B(\bA)$ is the product of the normalization constant for each Dirichlet distribution. The unnormalized posterior becomes
\begin{align*}
\mathscr{P} &\propto p(\bY, \bT \given \widehat{\bS}, \bX) \\[1mm]
\mathscr{P} &= p(\widehat{\bS} \given \bY, \bT, \bX) p(\bY, \bT \given \bX)
= p(\widehat{\bS} \given \bY, \bT, \bX) p(\bY, \bT) 
= p(\bT) p(\bY) \prod_{is} p(s \given \bx_i)^{\widehat{\bS}_{is}} \\
&= \frac{1}{B(\bA)} \prod_{ys} \bT_{ys}^{\bA_{ys}} 
\prod_{iy} p(y)^{\bY_{iy}} 
\prod_{is} \big( \bY \bT \big)_{is}^{\widehat{\bS}_{is}}.
\end{align*}
For the second line, note that $\bT$ and $\bY$ is per definition independent of $\bX$.

The log-likelihood is
\begin{align*}
\log \mathscr{L} = \sum_{is} \widehat{\bS}_{is} \log \big( \bY \bT \big)_{is}.
\end{align*}
The log-priors are
\begin{align*}
\log p(\bY) &= \log \left( \prod_{iy} p(y)^{\bY_{iy}} \right) = \sum_{iy} \bY_{iy} \log p(y) \\	
\log p(\bT) &= \log \left( \frac{1}{B(\bA)} \prod_{ys} \bT_{ys}^{\bA_{ys}} \right) 
= \log \left( \frac{1}{B(\bA)}\right) + \sum_{ys} \bA_{ys} \log \bT_{ys} 
\end{align*}
Thus the log-posterior becomes
\begin{align*}
\log \mathscr{P} &= \sum_{is} \widehat{\bS}_{is} \log \big( \bY \bT \big)_{is} 
+ \sum_{iy} \bY_{iy} \log p(y) + \log \left( \frac{1}{B(\bA)}\right) + \sum_{ys} \bA_{ys} \log \bT_{ys} 
\end{align*}
When optimizing the posterior we can disregard any term that is constant with respect to the elements being optimized. We will be optimizing with respect to $\bT$ and $\bY$ and therefore define an alternative optimization function $\mathscr{O}$ as
\begin{align*}
\mathscr{O} &= \log \mathscr{P} - \log \left( \frac{1}{B(\bA)}\right) - \sum_{is} \widehat{\bS}_{is} \log \widehat{\bS}_{is}
\\[2mm]
&= \sum_{is} \widehat{\bS}_{is} \log \big( \bY \bT \big)_{is} 
- \sum_{is} \widehat{\bS}_{is} \log \widehat{\bS}_{is}
+ \sum_{iy} \bY_{iy} \log p(y) + \sum_{ys} \bA_{ys} \log \bT_{ys} \\[1mm]
&= \sum_{is} \widehat{\bS}_{is} \log \left( \frac{\big( \widehat{\bY} \bT \big)_{is}}{\widehat{\bS}_{is}} \right) 
+ \sum_{iy} \bY_{iy} \log p(y) 
+ \sum_{ys} \bA_{ys} \log \bT_{ys} \\[2mm]
&= - \mathcal{KL}\left( \widehat{\bS}_{is}, \widehat{\bY} \bT \right) + \sum_{iy} \bY_{iy} \log p(y) + \sum_{ys} \bA_{ys} \log \bT_{ys}. 
\end{align*}
Here $\mathcal{KL}(\cdot)$ denotes the Kullback-Leibler divergence. We can thus optimize $\widehat{\bY}$ and $\widehat{\bT}$ by maximizing the negative KL-divergence with added terms from the Dirichlet priors on $\widehat{\bT}$ and $\widehat{\bY}$, solving task \ref{task:infer_Y_T_from_S}. If we know $\bT$ then we can remove the last term in $\mathscr{O}$ and simply optimize for the optimal $\widehat{\bY}$ which solves task \ref{task:predict_infer_Y_from_T_S}.

\subsection{Expected Likelihood} \label{sec:limit_mean_likelihood}

Consider a set $\mathcal{X}$ of $n$ samples $\bx_i \in \mathcal{X}$ from some true distribution $\mathcal{P}$. Their related classes are collected in a one-hot-encoded matrix $\bY \in \{0, 1\}^{n \times m}$ where each row is a sample and the columns represent the $m$ classes. If $p(y \given \bx_i, \mathcal{M})$ is the probability of class $y$ at sample $\bx_i$ under model $\mathcal{M}$, then the likelihood of this model is
\begin{align*}
\mathscr{L} =  p(\bY \given \mathcal{X}, \mathcal{M}) = \prod_{iy} p(y \given \bx_i, \mathcal{M})^{\bY_{iy}}.
\end{align*}
Here the matrix $\bY$ is simply used to select the probability of the appropriate class for each sample. The log-likelihood is
\begin{align*}
\log \mathscr{L} = \sum_{iy} \bY_{iy} \log p(y \given \bx_i, \mathcal{M}).
\end{align*}

Now say we are able to sample $\bY$ for the same set of points. We collect $N$ samples of $\bY$ into a tensor $\underline{\bY} \in \{0, 1\}^{N \times n \times m}$. The mean log-likelihood of a model becomes
\begin{align*}
\mean_{N,\bY} \; \log \mathscr{L}
&= \frac{1}{N} \sum_{j}^N  \sum_{iy} \underline{\bY}_{jiy} \log p(y \given \bx_i, \mathcal{M})
= \sum_{iy} \left( \frac{1}{N} \sum_{j}^N \underline{\bY}_{jiy} \right) \log p(y \given \bx_i, \mathcal{M}).
\end{align*}

The inner sum counts the number of observations at $\bx_i$ where the class was $y$. When dividing by the number of observations, we compute the empirical frequency of class $y$ at $\bx_i$. In the limit of $N$ approaching infinity this quantity becomes 
\begin{align*}
\lim_{N \rightarrow \infty} \frac{1}{N} \sum_{j}^N \underline{\bY}_{jiy} 
= \E_{\mathcal{P}} \big[ \bY_{i y} \big]
= p(y \given \bx_i).
\end{align*}

Thus the mean log-likelihood becomes
\begin{align*}
\lim_{N \rightarrow \infty} \mean_{N,\bY} \; \log \mathscr{L}
&= \E_{\mathcal{P}} \Big[ \log \mathscr{L} \Big] 
= \sum_{iy} p(y \given \bx_i) \log p(y \given \bx_i, \mathcal{M}).
\end{align*}

Be exponentiation we have
\begin{align*}
\exp \left( \E_{\mathcal{P}} \Big[ \log \mathscr{L} \Big] \right) 
= \prod_{iy} p(y \given \bx_i, \mathcal{M})^{p(y \given \bx_i)}.
\end{align*}

Thus the exponentiation of the mean log-likelihood can be expressed as a product of the model probabilities to the power of the true probabilities. This quantity can be estimated from data by using the observed frequencies instead of the true probabilities. \\

By Jensen's inequality we have (exponentiation is a convex function)
\begin{align*}
\exp \left( \E_{\mathcal{P}} \Big[ \log \mathscr{L} \Big] \right) 
\leq \E_{\mathcal{P}} \Big[ \exp \left( \log \mathscr{L} \right) \Big] = \E_{\mathcal{P}} \Big[ \mathscr{L} \Big]
\end{align*}

Thus maximizing the exponentiation of the expected log-likelihood acts as a surrogate which may maximize the likelihood.

\subsection{Direct Estimation of Transitions $\bT$} \label{sec:supplement_est_T}

For observed class densities $\widehat{\bY}$ and selection $\bS_{\mathcal{D}}$, the aggregated mass transitioned from each class to each selection is
\begin{align*}
\bM = \widehat{\bY}^{\top} \bS_{\mathcal{D}}.
\end{align*}
The likelihood of the transition matrix now becomes
\begin{align*}
p(\widehat{\bY}, \bS_{\mathcal{D}} \given \bT) = \prod_{ys} \bT_{ys}^{\bM_{ys}}.
\end{align*}
With a Dirichlet prior on $\bT$, the posterior is
\begin{align*}
p(\bT \given \widehat{\bY}, \bS_{\mathcal{D}}) 
&= \frac{
	p(\widehat{\bY}, \bS_{\mathcal{D}} \given \bT) p(\bT)
}{p(\widehat{\bY}, \bS_{\mathcal{D}})} 
\propto
p(\widehat{\bY}, \bS_{\mathcal{D}} \given \bT) p(\bT)
\end{align*}
The log of the posterior's numerator is
\begin{align}
\log p(&\widehat{\bY}, \bS_{\mathcal{D}} \given \bT) p(\bT) 
= \log \left( \prod_{ys} \bT_{ys}^{\bM_{y s}} \frac{1}{B(\bA)} 
\prod_{ys} \bT_{ys}^{\bA_{ys}} \right) \nonumber \\
&= \sum_{ys} \bM_{y s} \log \bT_{ys} + \frac{1}{B(\bA)} + \sum_{ys} \bA_{ys} \log \bT_{ys}.
\end{align}
We want to optimize the posterior subject to the constraint that the rows of $\bT$ sums to one. The Lagrangian of this problem (ignoring constant terms) is
\begin{align*}
\mathcal{L} &= \sum_{ys} \bM_{y s} \log \bT_{ys} + \sum_{ys} \bA_{ys} \log \bT_{ys}	
+ \blambda^{\top} (\bT \bone - \bone),
\qquad \blambda_y \not= 0.
\end{align*}
The derivative of the Lagrangian with respect to $\bT$ is
\begin{align*}
\diff[\mathcal{L}]{\bT} 
&= \diff{\bT}\sum_{ys} \bM_{y s} \log \bT_{ys} + \diff{\bT}\sum_{ys} \bA_{ys} \log \bT_{ys} + \bLambda \\
&= \bM \oslash \bT + \bA \oslash \bT + \bLambda
= (\bM + \bA) \oslash \bT + \bLambda, 
\end{align*}
where $\bLambda = [\blambda \;\; \blambda \;\; ... \;\; \blambda]$ is a matrix created by stacking $\blambda$ as column vectors $m_s$ times and $\oslash$ is elementwise division.
Setting this to zero we find that 
\begin{align}
\bT = -(\bM + \bA) \oslash \bLambda. \label{eq:T_from_M_A_Lambda}
\end{align}
We insert this into the constraint that $\bT \bone = \bone$ 
\begin{align}
\bone &= -\left( (\bM + \bA) \oslash \bLambda \right) \bone \nonumber 
= -(\bM + \bA) \bone \oslash \blambda \nonumber \\
\blambda &= - (\bM + \bA) \bone, \label{eq:lambda_solved}
\end{align}
so $\blambda$ is the negative row sums of $\bM +\bA$, and $\bT$ is the row-normalized version of $\bM + \bA$. In other words each element of $\bT$ is 
\begin{align*}
\bT_{ys} = \frac{(\bM + \bA)_{ys}}{\sum_s (\bM + \bA)_{ys}}.
\end{align*}
For $\bT$ to be estimated probabilities we require the rows to sum to one, as constrained using $\blambda$, but we also require non-negative elements. Since all elements of $\bM$ and $\bA$ are positive, this constraint is naturally satisfied. When optimizing $\bY$ and $\bT$, we can compute $\bT$ directly from this approach and thus only need to effectively optimize $\bY$ while always considering the optimal $\bT$.

\clearpage
\subsection{Costs}
Inspired by the work of \cite{du_plessis_analysis_2014} on cost-sensitive learning for positive-unlabelled learning, we here determine suitable costs for samples for the inference problem.

\subsubsection{Cost Sensitive Learning}
Say we have a cost-matrix $\bC$ where an element $\bC_{ss'}$ is the cost of assigning label $s'$ to a sample with actual label $s$. The cost of prediction using model $h(\bx)$ is a stochastic variable 
\begin{align*}
	C = \bC_{XY}, \qquad X \sim \mathcal{C}\big(p(s)\big), \qquad Y \sim \mathcal{C}\big( p(h(X) = s' \given s) \big),
\end{align*}
where $\mathcal{C}$ is a categorical distribution, $X$ is a label drawn from the prior distribution of labels and $Y$ is drawn from the predictive distribution of the model (evaluated on $X$).

The expected cost of prediction is
\begin{align}
	\E[C] &= \sum_{ss'} p(s) \: p(h(X) = s' \given s) \: \bC_{ss'} 
		= \sum_{ss'} p(s) \: R_{ss'} \: \bC_{ss'}, \label{eq:cost_sensitive_classification_form} \\
	R_{ss'} &= p(g(X) = s' \given s), \nonumber
\end{align}
where $p(g(X) = s' \given s)$ is the "risk" of selecting label $s'$ if the real label was $s$ (they may be the same). A classifier which seeks to minimize cost will therefore attempt to minimize an expression like the one above. \\

\subsubsection{Costs for Selection Densities} \label{sec:cost_selection_density}
We now focus on the problem on this article. Say we have access to $p(y \given s)$. In general $p(y \given s) \not= p(y \given s, \bx)$, but we now make a model $h(\bx)$ which estimates the class distribution by
\begin{align*}
	p(y \given \bx) 
		= \sum_s p(y \given s, \bx) \; p(s \given \bx) 
		\approx \sum_s p(y \given s) \; g_s(\bx) = h(\bx).
\end{align*}
The error rate of model $h(\bx)$ is
\begin{align*}
	R = \sum_y p(y) \sum_s p(s \given y) \sum_{s'} p(g(X) = s' \given s) \: p(h(X) \not= y \given s').
\end{align*}
The error produced directly by the decisions of $h$ is
\begin{align*}
	p(h(X) \not= y \given s') = 1 - p(y \given s').
\end{align*}
We therefore have
\begin{align}
	R 
		&= \sum_y p(y) \sum_s p(s \given y) \sum_{s'} p(g(X) = s' \given s) \: (1 - p(y \given s')) \nonumber \\
		&= \sum_{ss'} p(g(X) = s' \given s) \sum_y (1 - p(y \given s')) \: p(s \given y) \: p(y) \nonumber \\
		&= \sum_{ss'} R_{ss'} \sum_y (1 - p(y \given s')) \: p(s, y) \nonumber \\
		&= \sum_{ss'} p(s) \: R_{ss'} \bC_{ss'} \label{eq:selection_cost_form} \\
	\bC_{ss'} 
		&= \frac{\sum_y (1 - p(y \given s')) \: p(s, y)}{p(s)} 
		= \sum_y (1 - p(y \given s')) \: p(y \given s). \label{eq:selection_costs}
\end{align}
Where the risk in (\ref{eq:selection_cost_form}) is in the same form as (\ref{eq:cost_sensitive_classification_form}). The costs in (\ref{eq:selection_costs}) can be thought of as the "disagreement" between $s$ and $s'$.
We can compute the costs using $\bUpsilon$ by
\begin{align*}
	\bC &= \bUpsilon (1 - \bUpsilon)^{\top}.
\end{align*}

\clearpage
\subsection{Relation to Elkan and Noto 2008}  \label{sec:relation_to_elkan_and_noto_2008}
We will here show the relation of our methods to the method described in \cite{elkan_learning_2008}. We will be using the inverse of a $2\times 2$ matrix
\[
\bA = \begin{bmatrix}
a & b \\ c & d
\end{bmatrix} \qquad \qquad
\bA^{-1} = \frac{1}{ad - bc} \begin{bmatrix}
d & -b \\ -c & a
\end{bmatrix}.
\]

\subsubsection{Elkan and Noto 2008}
In \cite{elkan_learning_2008} they prove the following.
Say we have a positive class and a negative class, and that the probability of incorrectly selecting a negative as a positive is 0. \\
The probability of a selection is
\begin{align*}
p(s = 1 \given \bx) &= p(s = 1 \given y = 1) p(y = 1 \given \bx) \\
&+ p(s = 1 \given y = 0) p(y = 0 \given \bx) \\
&= p(s = 1 \given y = 1) p(y = 1 \given \bx) \\
&+ 0 \cdot p(y = 0 \given \bx) \\
&= p(s = 1 \given y = 1) p(y = 1 \given \bx) \\
\end{align*}
So therefore
\begin{align}
p(y = 1 \given \bx) &= \frac{p(s = 1 \given \bx)}{\rho} \label{eq:elkan_main_lemma} \\
\rho = p(s &= 1 \given y = 1) \nonumber
\end{align}

\cite{elkan_learning_2008} further concludes that
\begin{align}
	p(s = 1 \given \bx) \leq \rho \label{eq:elkan_p(s)_ceiling}
\end{align} 
in order for the probabilities to remain well behaved after scaling.

\subsubsection{Transition Version}
Using our methodology the corresponding transition matrix $\bT$ is
\begin{align*}
\bT = \begin{bmatrix}
\rho & 1 - \rho \\
0 & 1
\end{bmatrix}  \qquad \qquad \rho = p(s = 1 \given y = 1).
\end{align*}

The inverse is found by
\begin{align*}
\frac{1}{ad - bc} = \frac{1}{\rho \cdot 1 - (1 - \rho) \cdot 0} = \frac{1}{\rho} 
\end{align*}\begin{align*}
\begin{bmatrix}
d & -b \\ -c & a
\end{bmatrix} 
= 
\begin{bmatrix}
1 & \rho - 1 \\ 0 & \rho
\end{bmatrix} 
\end{align*}\begin{align*}
\bT^{-1} = 
\frac{1}{\rho}
\begin{bmatrix}
1 & \rho - 1 \\ 0 & \rho
\end{bmatrix} = 
\begin{bmatrix}
\frac{1}{\rho} & \frac{\rho - 1}{\rho} \\ 0 & 1
\end{bmatrix}
\end{align*}

The distribution across classes for a sample $\bx$ is (transposed for ease of reading)
\begin{align*}
	\bY^{\top} 
		&= \big( \bS \bT^{-1} \big)^{\top}
		= 
		\left( \Big[ \: p( s= 1 \given \bx) \;\;\; p( s= 0 \given \bx) \: \Big]
		\begin{bmatrix} \frac{1}{\rho} & \frac{\rho - 1}{\rho} \\ 0 & 1 \end{bmatrix} \right)^{\top} \\
		&= 
		\begin{bmatrix}
			\frac{1}{\rho} \cdot p( s= 1 \given \bx) \\
			\frac{\rho - 1}{\rho} \cdot p( s= 1 \given \bx) + p( s= 0 \given \bx)
		\end{bmatrix}
		= 
		\begin{bmatrix}
		p(y = 1 \given \bx) \\
		p(y = 0 \given \bx)
		\end{bmatrix}.
\end{align*}
We see that the probability of $y=1$ is $\frac{1}{\rho} \cdot p( s= 1 \given \bx)$ like in \ref{eq:elkan_main_lemma}. The constraint of \ref{eq:elkan_p(s)_ceiling} comes naturally from this result, but can also be showed from $p(y = 0 \given \bx)$ together with the corresponding constraint on $p(s=0 \given \bx)$ by \\
\begin{minipage}{0.4\textwidth}
	\begin{align*}
	0 &\leq p(y = 0 \given \bx) \\
	0 &\leq p(s = 1 \given \bx) \cdot \frac{\rho - 1}{\rho} + p(s = 0 \given \bx) \\
	0 &\leq p(s = 1 \given \bx) \cdot \frac{\rho - 1}{\rho} + 1 - p(s = 1 \given \bx) \\
	0 &\leq p(s = 1 \given \bx) \cdot \left(\frac{\rho - 1}{\rho} - \frac{\rho}{\rho}\right) + 1 \\
	-1 &\leq -\frac{1}{\rho} p(s = 1 \given \bx) \\
	1 &\geq \frac{1}{\rho} p(s = 1 \given \bx) \\
	\rho &\geq p(s = 1 \given \bx) \\
	\end{align*}
\end{minipage} \hfill \begin{minipage}{0.4\textwidth}
	\begin{align*}
	0 &\leq p(y = 0 \given \bx) \\
	0 &\leq (1 - p(s = 0 \given \bx)) \cdot \frac{\rho - 1}{\rho} + p(s = 0 \given \bx) \\
	0 &\leq \frac{\rho - 1}{\rho} - \frac{\rho - 1}{\rho} p(s = 0 \given \bx) + p(s = 0 \given \bx) \\
	0 &\leq \frac{\rho - 1}{\rho} + \left(\frac{\rho}{\rho} - \frac{\rho - 1}{\rho}\right) p(s = 0 \given \bx) \\
	0 &\leq \frac{\rho - 1}{\rho} + \frac{1}{\rho} p(s = 0 \given \bx) \\
	0 &\leq \rho - 1 + p(s = 0 \given \bx) \\
	1 - \rho &\leq p(s = 0 \given \bx).
	\end{align*}
\end{minipage}

\clearpage
\subsection{Performance Using Kernel Density on Fashion MNIST} \label{sec:kernel_fashion}

\renewcommand{\temppp}{-2.3mm}
\renewcommand{\tempp}{10.635mm} 
\renewcommand{\temp}{0.265}
\renewcommand{\tempppp}{2mm}
\begin{figure}[h!]
	\centering
	\begin{subfigure}{0.75\textwidth}
		\centering
		\includegraphics[width=0.25\textwidth,trim={0mm 0mm 0mm 0mm},page=16,clip]{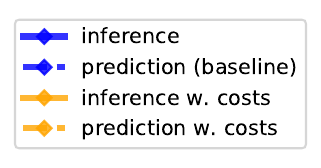} 
	\end{subfigure} \\
	\begin{subfigure}{0.5\textwidth}
		\centering
		\caption{Semi-Supervised, test score.} \vspace{\temppp}
		\includegraphics[scale=\temp,trim={0mm 0mm 0mm \tempp},page=3,clip]{figs/results/kernel_results_fashion.pdf} 
	\end{subfigure}%
	\begin{subfigure}{0.5\textwidth}
		\centering
		\caption{Semi-Supervised, training $\bW$-score.} \vspace{\temppp}
		\includegraphics[scale=\temp,trim={11mm 0mm 0mm \tempp},page=2,clip]{figs/results/kernel_results_fashion.pdf} 
	\end{subfigure} \\[\tempppp]
	\begin{subfigure}{0.5\textwidth}
		\centering
		\caption{7-Positive, test score.} \vspace{\temppp}
		\includegraphics[scale=\temp,trim={0mm 0mm 0mm \tempp},page=6,clip]{figs/results/kernel_results_fashion.pdf} 
	\end{subfigure}%
	\begin{subfigure}{0.5\textwidth}
		\centering
		\caption{7-Positive, training $\bW$-score.} \vspace{\temppp}
		\includegraphics[scale=\temp,trim={11mm 0mm 0mm \tempp},page=5,clip]{figs/results/kernel_results_fashion.pdf} 
	\end{subfigure} \\[\tempppp]
	\begin{subfigure}{0.5\textwidth}
		\centering
		\caption{Positive-Unlabelled, test score.} \vspace{\temppp}
		\includegraphics[scale=\temp,trim={0mm 0mm 0mm \tempp},page=9,clip]{figs/results/kernel_results_fashion.pdf} 
	\end{subfigure}%
	\begin{subfigure}{0.5\textwidth}
		\centering
		\caption{Positive-Unlabelled, training $\bW$-score.} \vspace{\temppp}
		\includegraphics[scale=\temp,trim={11mm 0mm 0mm \tempp},page=8,clip]{figs/results/kernel_results_fashion.pdf} 
	\end{subfigure} \\[\tempppp]
	\begin{subfigure}{0.5\textwidth}
		\centering
		\caption{Noisy-20, training $\bW$-score.} \vspace{\temppp}
		\includegraphics[scale=\temp,trim={0mm 0mm 0mm \tempp},page=11,clip]{figs/results/kernel_results_fashion.pdf} 
	\end{subfigure}%
	\begin{subfigure}{0.5\textwidth}
		\centering
		\caption{Noisy-50, training $\bW$-score.} \vspace{\temppp}
		\includegraphics[scale=\temp,trim={11mm 0mm 0mm \tempp},page=14,clip]{figs/results/kernel_results_fashion.pdf} 
	\end{subfigure} 
	\caption{Performance on Fashion MNIST dataset using kernel density estimation. We trained a kernel-density on the pixel values, after standardizing each feature to zero-mean and unit variance. When evaluating the kernel density on the training data we leave out the sample being evaluated. This allows density estimations that are less biased towards the training data. Using kernel-methods furthermore makes weighting of training samples trivial. \\
		Y-axis shows F1-score on all plots, while the X-axis shows the number of labelled instances in each (positive) class. The number of original samples in each class in the dataset is 6000, and so the graph ends at about 2/3 of the data being labelled. The blue curves uses flat cost on samples and the yellow curves use the costs from \ref{sec:costs}. The dashed lines are the baseline performance from predicting label with model and choosing the related class. The solid lines are the performance after inferring $\bT$ and $\bY$. Some of the plots show the $\bW$ predictions to illustrate performance on evaluating training samples. On the noise-label plot we have also included the performance gained from predicting on the unlabelled set and using the labels directly on the labelled instances (in black).}
\end{figure}

\clearpage
\subsection{Performance on 20 News Groups Dataset} \label{sec:20_news_groups}

\renewcommand{\temppp}{-2.3mm}
\renewcommand{\tempp}{10.63mm} 
\renewcommand{\temp}{0.265}
\renewcommand{\tempppp}{2mm}
\begin{figure}[!h]
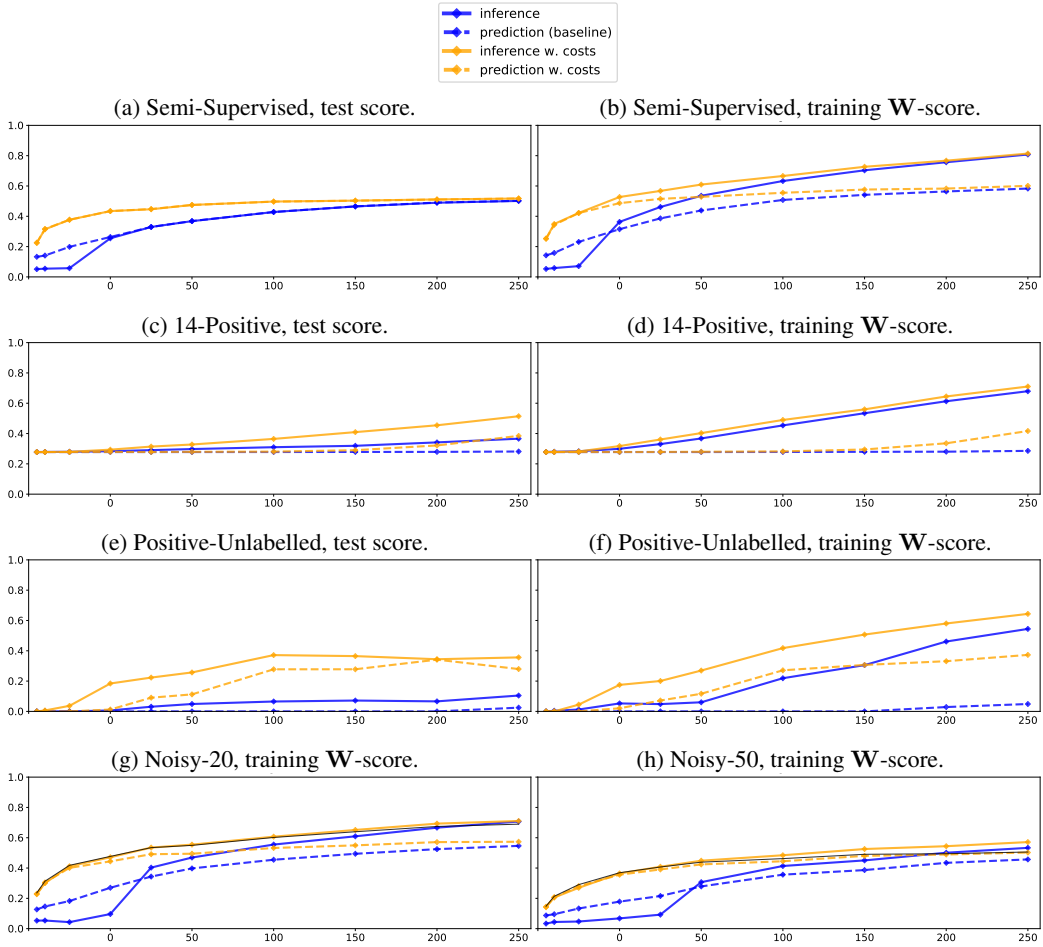

	\centering
	\begin{subfigure}{0.75\textwidth}
		\centering
		\includegraphics[width=0.25\textwidth,trim={0mm 0mm 0mm 0mm},page=19,clip]{figs/results/forest_results_20_news.pdf} 
	\end{subfigure} \\
	\begin{subfigure}{0.5\textwidth}
		\centering
		\caption{Semi-Supervised, test score.} \vspace{\temppp} 
		\includegraphics[scale=\temp,trim={0mm 0mm 0mm \tempp},page=3,clip]{figs/results/forest_results_20_news.pdf} 
	\end{subfigure}%
	\begin{subfigure}{0.5\textwidth}
		\centering
		\caption{Semi-Supervised, training $\bW$-score.} \vspace{\temppp}
		\includegraphics[scale=\temp,trim={11mm 0mm 0mm \tempp},page=2,clip]{figs/results/forest_results_20_news.pdf} 
	\end{subfigure} \\[\tempppp]
	\begin{subfigure}{0.5\textwidth}
		\centering
		\caption{14-Positive, test score.} \vspace{\temppp} 
		\includegraphics[scale=\temp,trim={0mm 0mm 0mm \tempp},page=6,clip]{figs/results/forest_results_20_news.pdf} 
	\end{subfigure}%
	\begin{subfigure}{0.5\textwidth}
		\centering
		\caption{14-Positive, training $\bW$-score.} \vspace{\temppp}
		\includegraphics[scale=\temp,trim={11mm 0mm 0mm \tempp},page=5,clip]{figs/results/forest_results_20_news.pdf} 
	\end{subfigure} \\[\tempppp]
	\begin{subfigure}{0.5\textwidth}
		\centering
		\caption{Positive-Unlabelled, test score.} \vspace{\temppp} 
		\includegraphics[scale=\temp,trim={0mm 0mm 0mm \tempp},page=12,clip]{figs/results/forest_results_20_news.pdf} 
	\end{subfigure}%
	\begin{subfigure}{0.5\textwidth}
		\centering
		\caption{Positive-Unlabelled, training $\bW$-score.} \vspace{\temppp} 
		\includegraphics[scale=\temp,trim={11mm 0mm 0mm \tempp},page=11,clip]{figs/results/forest_results_20_news.pdf} 
	\end{subfigure} \\[\tempppp]
	\begin{subfigure}{0.5\textwidth}
		\centering
		\caption{Noisy-20, training $\bW$-score.} \vspace{\temppp} 
		\includegraphics[scale=\temp,trim={0mm 0mm 0mm \tempp},page=14,clip]{figs/results/forest_results_20_news.pdf} 
	\end{subfigure}%
	\begin{subfigure}{0.5\textwidth}
		\centering
		\caption{Noisy-50, training $\bW$-score.} \vspace{\temppp} 
		\includegraphics[scale=\temp,trim={11mm 0mm 0mm \tempp},page=17,clip]{figs/results/forest_results_20_news.pdf} 
	\end{subfigure} 
	\caption{Performance on the 20 News Groups dataset using a random-forest classifier on word counts (stop-words removed) with our inference scheme (same method as in section \ref{sec:result_fashion}). Y-axis shows F1-score on all plots, while the X-axis shows the number of labelled instances in each (positive) class. The number of original samples in each class in the training set varies from 377 to 600. The blue curves uses flat cost on samples and the yellow curves use the costs from \ref{sec:costs}. The dashed lines are the baseline performance from predicting label with model and choosing the related class. The solid lines are the performance after inferring $\bT$ and $\bY$. Some of the plots show the $\bW$ predictions to illustrate performance on evaluating training samples. On the noise-label plot we have also included the performance gained from predicting on the unlabelled set and using the labels directly on the labelled instances (in black).} \label{fig:results_forest_20_news}
\end{figure}

\end{document}